\pdfoutput=1

\documentclass[11pt]{article}

\usepackage[]{EMNLP2022}

\usepackage{times}
\usepackage{latexsym}

\usepackage[T1]{fontenc}

\usepackage[utf8]{inputenc}

\usepackage{microtype}

\usepackage{inconsolata}

\usepackage{booktabs}
\usepackage{xcolor}         %
\usepackage{graphicx}
\usepackage{xspace}
\usepackage{todonotes}
\usepackage{multirow}
\usepackage{wrapfig}
\usepackage{caption}
\usepackage{subcaption}
\usepackage{tabularx}
\usepackage{amsmath}
\usepackage{amsfonts}
\usepackage{bbm}
\usepackage{bm}
\usepackage{mathtools}
\usepackage{lipsum}
\usepackage[ruled,vlined]{algorithm2e}

\usepackage{pifont}%

\newcommand{\eat}[1]{}

\newcommand{\datasize}[0]{\textsc{DataSize}\xspace}
\newcommand{\curvegrad}[0]{\textsc{CurveGrad}\xspace}
\newcommand{\textemb}[0]{\textsc{TextEmb}\xspace}
\newcommand{\taskemb}[0]{\textsc{TaskEmb}\xspace}
\newcommand{\prompt}[0]{\textsc{PTuning}\xspace}
\newcommand{\lora}[0]{\textsc{LoRA}\xspace}
\newcommand{\bitfit}[0]{\textsc{BitFit}\xspace}

\newcommand{\baby}{TuPaTE\xspace}

\title{Efficiently Tuned Parameters are Task Embeddings}

\author{Wangchunshu Zhou$^1$\thanks{\ \ Equal contribution.} , Canwen Xu$^{2*}$, Julian McAuley$^2$\\
 $^1$ ETH Zurich
 $^2$ University of California, San Diego \\
 $^1$\texttt{wangchunshu.zhou@inf.ethz.ch,} $^2$\texttt{\{cxu,jmcauley\}@ucsd.edu}  \\
 }

\begin{document}
\maketitle

\begin{abstract}
Intermediate-task transfer can
benefit a wide range of NLP tasks with properly selected source datasets. 
However, it is computationally infeasible to experiment with all intermediate transfer combinations, making choosing a useful source task a challenging problem. 
In this paper, we
anticipate
that task-specific parameters updated in parameter-efficient tuning methods are likely to encode task-specific information. Therefore, such parameters can be predictive for inter-task transferability.
Thus, we propose to exploit these efficiently tuned parameters as off-the-shelf task embeddings for the efficient selection of source datasets for intermediate-task transfer. 
We experiment with
11 text classification tasks and 11 question answering tasks. 
Experimental results show that our approach can consistently outperform existing inter-task transferability prediction methods while
being conceptually simple and computationally efficient. 
Our analysis also reveals that the ability of efficiently tuned parameters on transferability prediction is disentangled with their in-task performance. This allows us to use parameters from early checkpoints
as task embeddings to further improve efficiency.\footnote{Code available at \url{https://github.com/JetRunner/TuPaTE}.}
\end{abstract}

\section{Introduction}

The \textit{pretraining then fine-tuning} paradigm~\citep{peters2018elmo,devlin2018bert,radford2018improving,radford2019language,brown2020language,lewis2019bart,raffel2019exploring} has substantially improved the state-of-the-art on a wide range of natural language processing (NLP) tasks. In this paradigm, we first pretrain a large language model on large-scale corpora in a general domain, and then fine-tune the pretrained model to be a task-specific model on the target dataset. In addition to directly transferring from a general pretrained language model, prior work~\citep{phang2018sentence} also shows that \textit{intermediate-task transfer}, i.e., fine-tuning on intermediate source tasks before the target task, can further improve target task performance. However, the success of intermediate-task transfer heavily relies on the selection of a proper source dataset while an inappropriate source dataset often leads to performance degradation compared to plain fine-tuning. Therefore, some recent works~\citep{vu2020exploring,poth2021pre} investigate methods to efficiently predict inter-task transferability without actually trying out all intermediate-task combinations.

\begin{figure}
    \centering
    \includegraphics[width=\columnwidth]{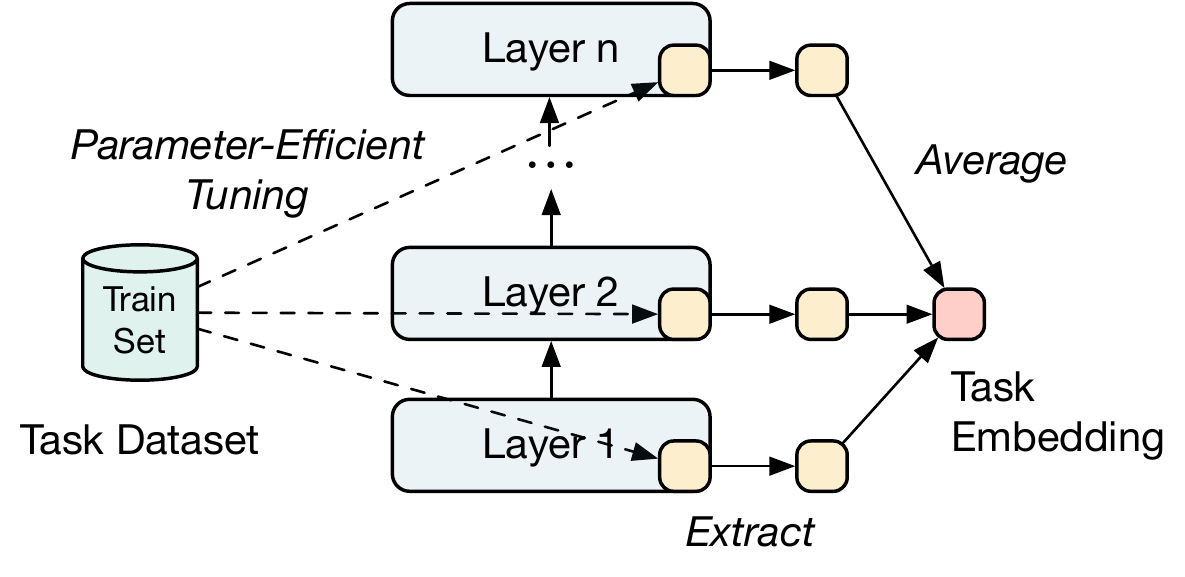}
    \caption{The workflow of using efficiently tuned parameters as task embeddings. The yellow boxes represent tunable parameters in Transformer layers.}
    \vspace{-0.5cm}
    \label{fig:workflow}
\end{figure}

The current state of the art~\citep{vu2020exploring} on predicting inter-task transferability is built on Task2Vec~\citep{achille2019task}, which considers the Fisher information matrix of a model fine-tuned on a task as the ``task embedding'', and predicts inter-task transferability by computing the cosine similarity between the task embedding of the source and target tasks. Despite empirically performing well, this approach requires fine-tuning the full model and (inefficiently) computing the Fisher matrix of the model.
Moreover, the resulting task embeddings generally have a high dimensionality similar to the size of the underlying model. Therefore, intermediate task selection, which requires storing task embeddings for each source/target task, can be %
space-consuming, especially when experimenting with large language models.

In this work, we opt for parameter-efficient tuning approaches~\citep{houlsby2019parameter,li2021prefix,guo2020parameter, hu2021lora,bitfit} for the efficient and accurate prediction of inter-task transferability. Our key insight is that task-specific parameters updated in parameter-efficient tuning methods are likely to encode high density task-specific information since they are used as a query for retrieving task-related knowledge in a frozen pretrained language model. Therefore, we propose to directly use task-specific parameters learned via parameter-efficient tuning on source/target datasets as task embeddings, as shown in Figure~\ref{fig:workflow}. Compared to task embeddings obtained by calculating the Fisher matrix of the fine-tuned model~\citep{achille2019task,vu2020exploring}, efficiently tuned parameters are of much lower dimensionality and do not suffer from noise from uninformative weights in the model parameters, thus leading to more accurate transferability prediction. Also, our method only requires parameter-efficient tuning on the tasks and stores task-specific parameters, making both computing and storing task embeddings more efficient. Moreover, with the development of open-source parameter-efficient tuning platforms like AdapterHub~\citep{pfeiffer2020adapterhub}, we can easily access off-the-shelf %
parameters of the source and target datasets downloaded from the model zoo and then compute the %
similarity between the downloaded parameters.

We empirically verify the effectiveness of our approach by experimenting with
11 text classification tasks and 11 question answering tasks, following~\citet{vu2020exploring}. Our results show that our approach consistently outperforms existing inter-task transferability prediction methods while being simpler and more efficient. In addition, we find that the
ability of efficiently tuned parameters on transferability prediction is not strongly correlated with their in-task performance. Therefore, task-specific parameters tuned with a relatively small number of steps are already highly predictive for inter-task transferability, allowing us to further improve the efficiency of intermediate task selection.

\section{Related Work}

Prior work~\citep{phang2018sentence} shows that positive transfer can be elicited by training a model on intermediate source tasks before fine-tuning on the target task. However, the choice of an appropriate source task is crucial for effective transfer. \citet{phang2018sentence} show that the size of the source dataset is an good prior for source task selection. \citet{pruksachatkun2020intermediate} propose to use task requiring complex reasoning and inference as source tasks. Besides these heuristics, a number of work also focuses on systematic prediction of intermediate task transferability. \citet{vu2020exploring} propose to used \textsc{Task2Vec} to construct task embeddings based on the input text or Fisher information matrix of a fine-tuned model. \citet{poth2021pre} further extend similar ideas for adapter-based transfer learning. More recently, \citet{vu2021spot} explore prompt-based transfer and propose to use prompt similarity as a predictor for prompt transferability to select proper soft prompts for initialization. This can be viewed as a special case of our proposed method where the parameter-efficient tuning method is restricted to vanilla prompt tuning~\citep{lester2021power} and the transfer method is restricted to prompt transfer instead of general intermediate-task transfer.
\section{Methodology}
\subsection{Parameter-Efficient Tuning} 
Parameter-efficient tuning only updates a small portion of parameters in a large pretrained model. In this paper, we experiment with three types of parameter-efficient tuning: Prompt Tuning~\citep{liu2021p}, Bias Tuning~\citep{bitfit}, and Low-Rank Tuning~\citep{hu2021lora}.

\paragraph{Prompt Tuning} We experiment with P-Tuning v2~\citep{liu2021p}. Specifically, P-Tuning v2 implements a prompt tuning method by introducing additional attention prefix matrices $K_t = \{\mathbf{k}_1 \ldots \mathbf{k}_n\} $ and $V_t = \{\mathbf{v}_1 \ldots \mathbf{v}_n\}$ for each Transformer layer, where $n$ is a hyperparameter controlling the added prefix length; $\mathbf{k}_*$ and $\mathbf{v}_*$ are vectors with dimension $d_h$; $d_h$ is the hidden size of the Transformer model.

For each Transformer layer, the added vectors are concatenated with the original key and value matrices to be $K' = K_t \oplus K$ and $V' = V_t \oplus V$, where $K$ and $V$ are the original key and value in each layer's attention block.
Then, the new scaled dot-product attention is calculated by replacing the original $K$ and $V$ with the new $K'$ and $V'$, respectively.

\paragraph{Bias Tuning} BitFit~\citep{bitfit} simply updates all bias terms $b$ in all linear layers $h=Wx + b$ in each Transformer layer.

\paragraph{Low-Rank Tuning}
LoRA~\cite{hu2021lora} injects trainable rank decomposition matrices into each layer of the Transformer model. For each linear layer $h=Wx$ where $W \in \mathbb{R}^{d \times k}$, the forward pass is modified to $h=Wx+BAx$, where $B \in \mathbb{R}^{d\times r}$, $A \in \mathbb{R}^{r \times k}$, and the rank $r \ll min(d,k)$.

\subsection{Tuned Parameters as Task Embeddings}
After parameter-efficient tuning, we concatenate all tuned parameters in each Transformer layer and average them across all layers to obtain a vector as a representation for a task, namely \textbf{Tu}ned \textbf{P}arameters \textbf{a}s \textbf{T}ask \textbf{E}mbedding (TuPaTE). Following \citet{vu2020exploring}, we calculate the cosine similarity between the embeddings of a given targeted task and the candidate source tasks. Then, we rank the candidate source tasks in descending order by the similarity scores.

\section{Experiments}
\subsection{Datasets}

Following \citet{vu2020exploring}, we conduct experiments with 11 tasks of text classification or regression (CR) and 11 tasks of question answering (QA). Note that \citet{vu2020exploring} also includes 11 tasks of sequence labeling. We do not include those datasets since most of them are not publicly available. The list of datasets can be found in Appendix~\ref{sec:appendix}. To be consistent with \citet{vu2020exploring}, we use two metrics to evaluate the performance of the task embeddings: (1) the average rank $\rho$ of the source task with the \textit{highest} absolute transfer gain; (2) Normalized Discounted Cumulative Gain (NDCG), which is a widely used metric for evaluating the quality of the entire ranking, instead of focusing on the highest rank as $\rho$ does. 

\subsection{Baselines}
We use the following methods as baselines: (1) \textbf{\datasize}~\citep{vu2020exploring} is a simple baseline that ranks all source tasks by the number of training examples. (2) \textbf{\curvegrad}~\citep{bingel2017eacl,vu2020exploring} is a baseline that uses the gradients of the loss curve of BERT for each task. It is originally proposed in \citet{bingel2017eacl} for predicting gains from multi-task learning and adapted by \citet{vu2020exploring} for predicting transferability. (3) \textbf{\textemb}~\citep{vu2020exploring} averages sentence representations over the entire dataset. The sentence representation is obtained by averaging the hidden states in the last layer of BERT. (4) \textbf{\taskemb}~\citep{vu2020exploring} represents tasks based on the Fisher information matrix. It is adapted from the task embedding originally proposed in \citet{achille2019task} for meta-learning.

\subsection{Training Details}

We apply P-Tuning v2, BitFit, and LoRA on BERT-base for fine-tuning on the aforementioned datasets. For each method, we adopt the default hyperparameters from their corresponding papers. Specifically, for P-Tuning v2, we use a prefix length of 20 and search the learning rate from \{1e-2, 1e-3\}; For LoRA, we set LoRA's $r$ to 8 and $\alpha$ to 8, and search a learning rate from \{5e-4, 2e-4\}; For BitFit, we search a learning rate from \{1e-4, 4e-4\}. We train all models with a batch size of 32 for 20 epochs on all datasets. We use the parameters tuned for 2 epochs as ``early'' task embeddings and those corresponding to the best validation set performance as ``late'' task embeddings. We compare the number of tunable parameters and the final task embedding dimensions in Table \ref{tab:method}. We can see that \baby has a significantly lower dimensionality compare to the \taskemb baseline. We also include an ensemble of the three efficient tuning methods (denoted as ``\textsc{3 Ensemble}''), by averaging the inter-task similarity scores of each model.

\begin{table}[t!]
    \centering
    \begin{tabular}{lrr}
    \toprule
    \multirow{2}{*}{Method} & \#Tuned & Embedding \\
    & Param. & Dim. \\
    \midrule
    \taskemb & 110M & 110M   \\
    \midrule
        \prompt & 184K & 15.4K   \\
        \lora & 300K & 25.0K \\
        \bitfit & 100K & 8.3K \\
    \bottomrule
    \end{tabular}
    \caption{Numbers of tuned parameters and the dimensions of the final task representation. }
    \label{tab:method}
\end{table}

\begin{table*}[t!]
\centering
\resizebox{\textwidth}{!}{
\begin{tabular}{l l rlrl rlrl rlrl}
\toprule
\multirow{4}{*}{Task Type} & \multirow{4}{*}{Method} & \multicolumn{4}{c}{\textsc{Full $\rightarrow$ Full}} &
\multicolumn{4}{c}{\textsc{Full $\rightarrow$ Limited}} &
\multicolumn{4}{c}{\textsc{Limited $\rightarrow$ Limited}}\\
\cmidrule(l){3-6} \cmidrule(l){7-10} \cmidrule(l){11-14}
& & \multicolumn{2}{c}{\emph{in-class (10)}} & \multicolumn{2}{c}{\emph{all-class (21)}} & \multicolumn{2}{c}{\emph{in-class (10)}} & \multicolumn{2}{c}{\emph{all-class (21)}} & \multicolumn{2}{c}{\emph{in-class (10)}} & \multicolumn{2}{c}{\emph{all-class (21)}}\\
\cmidrule(lr){3-4} \cmidrule(lr){5-6} 
\cmidrule(lr){7-8} \cmidrule(lr){9-10} 
\cmidrule(lr){11-12} \cmidrule(lr){13-14}  
& & \multicolumn{1}{l}{$\rho\downarrow$} & \multicolumn{1}{l}{NDCG$\uparrow$} & \multicolumn{1}{l}{$\rho\downarrow$} & \multicolumn{1}{l}{NDCG$\uparrow$} &\multicolumn{1}{l}{$\rho\downarrow$} & \multicolumn{1}{l}{NDCG$\uparrow$} &\multicolumn{1}{l}{$\rho\downarrow$} & \multicolumn{1}{l}{NDCG$\uparrow$} &\multicolumn{1}{l}{$\rho\downarrow$} & \multicolumn{1}{l}{NDCG$\uparrow$} & \multicolumn{1}{l}{$\rho\downarrow$} & \multicolumn{1}{l}{NDCG$\uparrow$} \\
\midrule 
\multirow{8}{*}{Classification/} & \datasize &3.6	&80.4	& 7.8	& 75.2	
&3.8	&62.9	& 8.9	& 57.2 &
\multicolumn{1}{c}{-}&\multicolumn{1}{c}{-}&\multicolumn{1}{c}{-}&\multicolumn{1}{c}{-}\\
\multirow{8}{*}{Regression (CR)} & \curvegrad 	&5.5	&68.6	& \multicolumn{1}{c}{-}  & \multicolumn{1}{c}{-}  & 6.4	&45.2	& \multicolumn{1}{c}{-}	& \multicolumn{1}{c}{-}	&5.9	&50.8	& \multicolumn{1}{c}{-} & \multicolumn{1}{c}{-} \\
& \textemb	& 5.2	& 76.4	& 9.8	& 74.7
&3.5	&60.3	& 7.5	& 55.6
&4.8	&61.4	& 11.4	& 46.2 \\
& \taskemb	&2.8	& 82.3	& 5.4	& 78.3
& 3.4	& 68.2	& 7.1	& 63.5
& 4.2	&62.6	& 9.7	&  47.7 \\
\cmidrule{2-14}
& \textsc{\baby} \\
& \ +\prompt	& 2.5	& 83.7	&  4.5 &   81.0
& \bf 3.1	&  71.3	& 6.4 &  65.1
&  3.9	& 64.6	&  8.1	& 51.3 \\
& \ +\lora & 2.7	& 83.0	&   5.0 &  79.6
& 3.3	& 70.5	&   6.8 &   63.7
& 4.0	& 64.2	&   9.0	&   49.3 \\
& \ +\bitfit	&  2.5	& 83.5	&   4.3 &  81.6
& 3.2	& 71.1	&  6.5 &   64.6
& \bf 3.8	& 64.9	&   8.3	&  50.9 \\
& \ 3 \textsc{Ensemble}  & \bf 2.3	& \bf 83.9	&  \bf 4.2 &  \bf 81.8
& \bf 3.1	&  \bf 71.5	&  \bf 6.2 & \bf 65.3
& \bf 3.8	& \bf 65.1	&   \bf 8.0	&  \bf 51.5 \\
\midrule 
\multirow{8}{*}{Question} & \datasize	& 3.2	& 84.4	& 11.4	& 65.8
&2.3	& 77.0	& 11.2	& 43.5
&\multicolumn{1}{c}{-}&\multicolumn{1}{c}{-}&\multicolumn{1}{c}{-}&\multicolumn{1}{c}{-}\\	
\multirow{8}{*}{Answering (QA)} & \curvegrad 	&8.3	&64.8	&\multicolumn{1}{c}{-}	&\multicolumn{1}{c}{-}	&8.2	&49.1	&\multicolumn{1}{c}{-}	&\multicolumn{1}{c}{-}	&6.8	&53.4	&\multicolumn{1}{c}{-}	&\multicolumn{1}{c}{-}\\
& \textemb	&4.1	& 81.1	& 5.8	& 82.0
&2.7	&77.6	& 3.8	& 80.5
&4.1	&65.6	&7.3	& 69.1\\
& \taskemb	& 3.2	&84.5	& 5.4	& 82.8
&2.5	&78.0	& 3.6	& 81.6
& 3.6	& 67.1	&7.1	& 69.5\\
\cmidrule{2-14}
& \textsc{\baby} \\
& \ +\prompt & 3.0	& 85.7	& \textbf{4.8}	& 83.3
& 2.2	& 80.9	& 3.1 & 83.5
& \bf 3.2	&  \bf 68.3	& \textbf{6.3}  & \textbf{72.4} \\
& \ +\lora & 3.1	& 85.3	& 5.2	& 83.0
& 2.3	& 79.8	& 3.3 & 82.5
& 3.4	&  67.5	& 6.7  & 70.8 \\
& \ +\bitfit &  3.0	& 85.5	& 4.9	& 83.1
&  2.1	& 81.4	& 3.1 & 83.4
& 3.3	&  68.0	& 6.5  & 72.0 \\
& \ 3 \textsc{Ensemble} & \bf 2.9	& \bf 85.9	& \bf 4.8	& \bf 83.5
& \bf 2.0	& \bf 81.7	& \bf 2.9 & \bf 83.7
& \bf 3.2	& 68.2	&  \bf 6.3  & \bf 72.4 \\
\bottomrule
\end{tabular}}
\caption{To evaluate \baby, we measure the average rank ($\rho$) %
assigned to the best source task (i.e., the one that results in the largest transfer gain) across target tasks, as well as the average NDCG measure of the overall ranking's quality. Parentheses denote the number of source tasks in each setting. Some results of \curvegrad are missing (marked with ``-'') since its code is not available. The other results of \curvegrad are taken from \citet{vu2020exploring}.  }
\label{table:taskemb}
\vspace{-1.5mm}
\end{table*}

\subsection{Experimental Results}

\begin{table}[t]
    \centering
    \resizebox{0.9\columnwidth}{!}{
    \begin{tabular}{lrrl}
    \toprule
    \multirow{2}{*}{Method} & \multirow{2}{*}{$\Delta\rho$} & \multirow{2}{*}{$\Delta$NDCG} &
    NCDG-Perf. \\
    & & & Pearson \\
    \midrule
        \prompt & 0.0 & $+$0.1 & 0.25  \\
        \lora & 0.0 & $-$0.1 & 0.17 \\
        \bitfit & $+$0.1 & $+$0.2 & 0.20 \\
    \bottomrule
    \end{tabular}
    }
    \caption{Analysis on the correlation between task-specific performance (e.g., accuracy) and transferability prediction results (i.e., $\rho$ and NDCG) for different parameter-efficient tuning methods. $\Delta\rho$ and $\Delta$NDCG denote the difference of $\rho$ and NDCG between the parameters with the highest and lowest task-specific performance. }
    \label{tab:correlation}
\end{table}

\begin{table}[t]
    \centering
    \resizebox{0.85\columnwidth}{!}{
    \begin{tabular}{lrlrl}
    \toprule
    \multirow{2}{*}{Method} & \multicolumn{2}{c}{Early} & \multicolumn{2}{c}{Best} \\
    \cmidrule(l){2-3} \cmidrule(l){4-5}
    & $\rho$ & NDCG & $\rho$ & NDCG \\
    \midrule
        \prompt & 2.5 & 83.5 & 2.5 & 83.7  \\
        \lora & 2.8 & 82.6 & 2.7 & 83.0 \\
        \bitfit & 2.5 & 83.2 & 2.5 & 83.5 \\
    \bottomrule
    \end{tabular}
    }
    \caption{Transferability prediction results with early checkpoints (checkpoints after 2 epochs) and the best checkpoints (checkpoints corresponding to the best validation performance). }
    \label{tab:early}
\end{table}

We present the main results in Table \ref{table:taskemb}. We find that \baby with different parameter-efficient tuning methods consistently outperforms prior works including \textemb\ and \taskemb. Interestingly, the performance improvement is larger in \textsc{Full $\rightarrow$ Limited} and \textsc{Limited $\rightarrow$ Limited} settings. We conjecture that this is because in limited resource settings, parameter-efficient tuning methods generally perform much better than full fine-tuning, which is used in the \taskemb method. Moreover, we find that \prompt and \bitfit outperform \lora in all settings. We suspect this is because the amount of tunable parameters in \lora is much larger than \prompt and \bitfit. Also, the ensemble of three methods achieve even better performance than only using one approach. 

\subsection{Analysis}

We conduct additional experiments in the \textit{in-class} setting on classification/regression tasks to better understand how \baby works. We first analyze the correlation between the in-task performance (e.g., accuracy) and transferability prediction ability of efficiently tuned parameters. We train \baby with 5 random combinations between searchable hyperparameters and random seeds, and present the correlation in Table \ref{tab:correlation}. We observe that there is only a weak correlation between 
in-task performance and transferability prediction results, indicating that the ability of efficiently tuned parameters to encode task-related information is disentangled with their final in-task performance. This also shows the robustness of \baby with respect to hyperparameters.

The fact that in-task performance only correlates weakly with transferability prediction motivates us to explore whether early checkpoints of efficiently tuned parameters can be used for transferability prediction. From Table~\ref{tab:early}, we find that early checkpoints are also effective task embeddings. This allows us to reduce the computation cost by around 90\% while substantially outperforming the \taskemb\ baseline.

\section{Conclusion}
In this paper, we show that efficiently tuned parameters are highly predictive for inter-task transferability and thus can be used as off-the-shelf task embeddings for source task selection in intermediate-task transfer learning. Our empirical investigation with three parameter-efficient tuning methods on 22 NLP tasks demonstrates that our approach outperforms prior works on inter-task transferability prediction despite being more efficient.

\section*{Limitations}
We select three representative works for three types of parameter-efficient tuning. However, there are other parameter-efficient tuning methods that we have not investigated. Although we believe our conclusion can generalize to other methods, we will conduct more experiments to confirm for future work.

\section*{Ethics Statement}
We propose to use efficiently tuned parameters as task embedding, only for predicting the performance of intermediate transfer learning. Thus, we do not anticipate any major ethical concern.

\section*{Acknowledgements}
We would like to thank the anonymous reviewers for their insightful comments. This project is partly supported by NSF Award \#1750063.

\bibliography{anthology,custom}
\bibliographystyle{acl_natbib}

\clearpage
\appendix

\section{List of Datasets}
\label{sec:appendix}

\begin{table}[h!]
\centering
\begin{tabular}{lr}
\toprule
\multicolumn{1}{l}{\textbf{Task}} & \multicolumn{1}{r}{\textbf{$|$Train$|$}} \\
\midrule
\multicolumn{2}{l}{\emph{Text classification / Regression (CR)}} \\
\cmidrule{1-1}
SNLI~\cite{SBowman15} & 570k \\
MNLI~\cite{AWilliams18} & 393k \\
QQP~\cite{Quora17} & 364k \\
QNLI~\cite{glue} & 105k \\
SST-2~\cite{RSocher13} &  67k \\
SciTail~\cite{TKhot18} & 27k \\
CoLA~\cite{AWarstadt19} &  8.5k \\
STS-B~\cite{DCer17} & 7k \\
MRPC~\cite{WDolan05} & 3.7k \\
RTE~\cite{IDagan06} & 2.5k \\
WNLI~\cite{HLevesque11} & 634 \\
\midrule
\multicolumn{2}{l}{\emph{Question Answering (QA)}} \\
\cmidrule{1-1}
SQuAD-2~\cite{PRajpurkar18} & 162k \\
NewsQA~\cite{ATrischler17} & 120k \\
HotpotQA~\cite{ZYang18} & 113k \\
SQuAD-1~\cite{PRajpurkar16} & 108k \\
DuoRC-p~\cite{ASaha18} & 100k \\
DuoRC-s~\cite{ASaha18} & 86k \\
DROP~\cite{DDua19} & 77k \\
WikiHop~\cite{JWelbl18} & 51k \\
BoolQ~\cite{CClark19} & 16k \\
ComQA~\cite{AAbujabal19} & 11k \\
CQ~\cite{JBao16} & 2k \\
\bottomrule
\end{tabular}
\caption{The datasets used in our experiments and their training set size \cite{vu2020exploring}.}
\label{tab:datasets}
\vspace*{-2mm}
\end{table}

\end{document}